\documentclass[11pt,a4paper]{article}
\usepackage[hyperref]{acl2017}
\usepackage{times}
\usepackage{latexsym}
\usepackage{amsmath}
\usepackage{amsfonts}
\usepackage{bm}
\usepackage{url}
\usepackage{booktabs}
\usepackage{tabularx}
\usepackage{multirow}
\usepackage{graphicx}
\usepackage{subcaption}

\graphicspath{ {figures/} }

\DeclareMathOperator*{\argmax}{argmax}

\aclfinalcopy % Uncomment this line for the final submission

\title{Density Estimation for Geolocation via Convolutional Mixture Density Network}

\author{Hayate Iso, Shoko Wakamiya, Eiji Aramaki \\ Nara Institute of Science and Technology \\  {\tt \{iso.hayate.id3,wakamiya,aramaki\}@is.naist.jp}}

\date{}

\begin{document}
\maketitle
\begin{abstract}
Nowadays, geographic information related to Twitter is crucially important for fine-grained applications. However, the amount of geographic information avail- able on Twitter is low, which makes the pursuit of many applications challenging. Under such circumstances, estimating the location of a tweet is an important goal of the study. Unlike most previous studies that estimate the pre-defined district as the classification task, this study employs a probability distribution to represent richer information of the tweet, not only the location but also its ambiguity. To realize this modeling, we propose the convolutional mixture density network (CMDN), which uses text data to estimate the mixture model parameters. Experimentally obtained results reveal that CMDN achieved the highest prediction performance among the method for predicting the exact coordinates. It also provides a quantitative representation of the location ambiguity for each tweet that properly works for extracting the reliable location estimations.

\end{abstract}

\begin{figure*}[t]
      \centering
      \begin{subfigure}[b]{0.319\textwidth}
      \includegraphics[width=\textwidth]{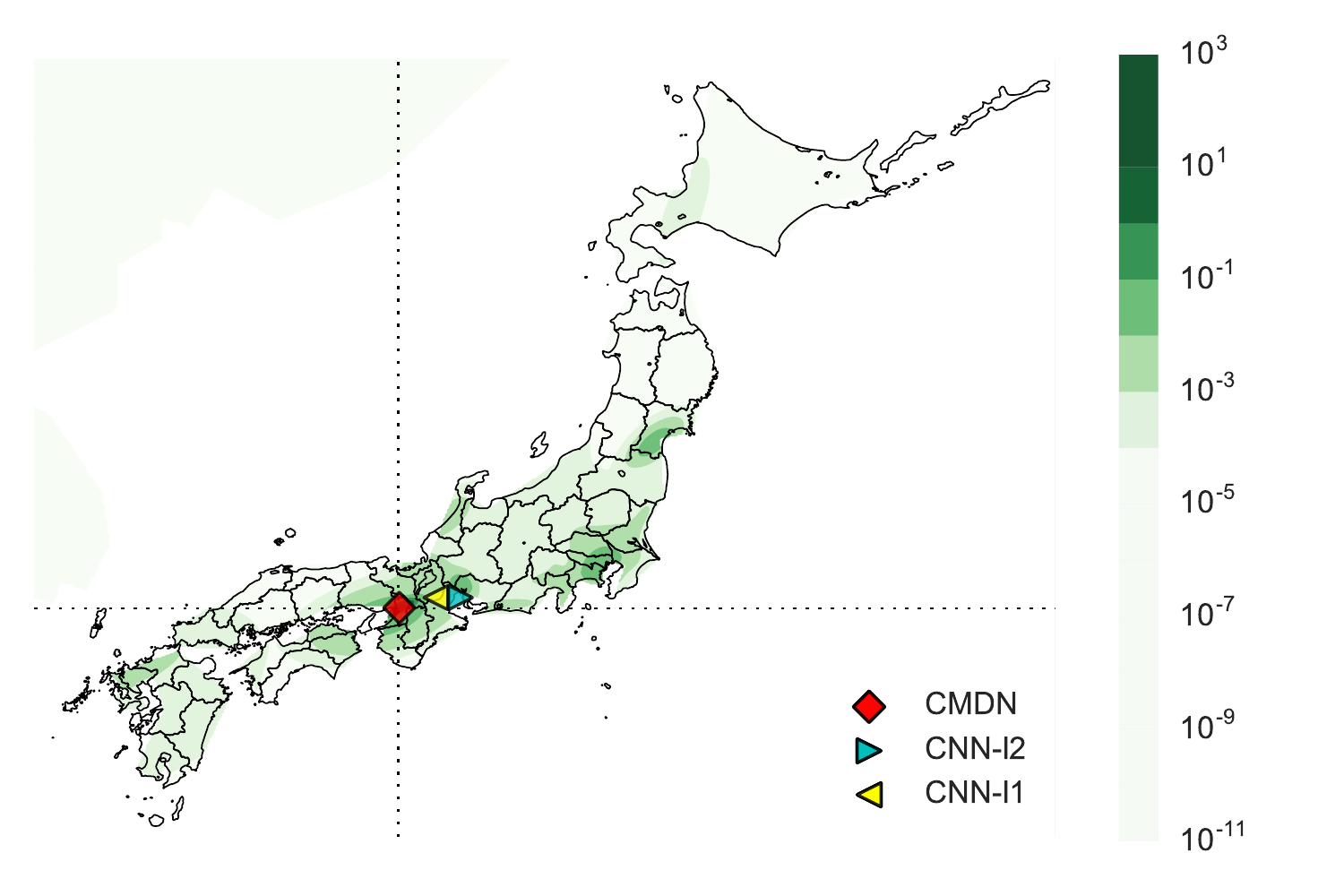}
        \caption{\textit{I got lost at Umeda station.. Here is almost like a jungle...}\\ likelihood:195.5, distance: 2.15 (km)}
        \label{fig:1a}
      \end{subfigure}
      ~
      \begin{subfigure}[b]{0.319\textwidth}
        \includegraphics[width=\textwidth]{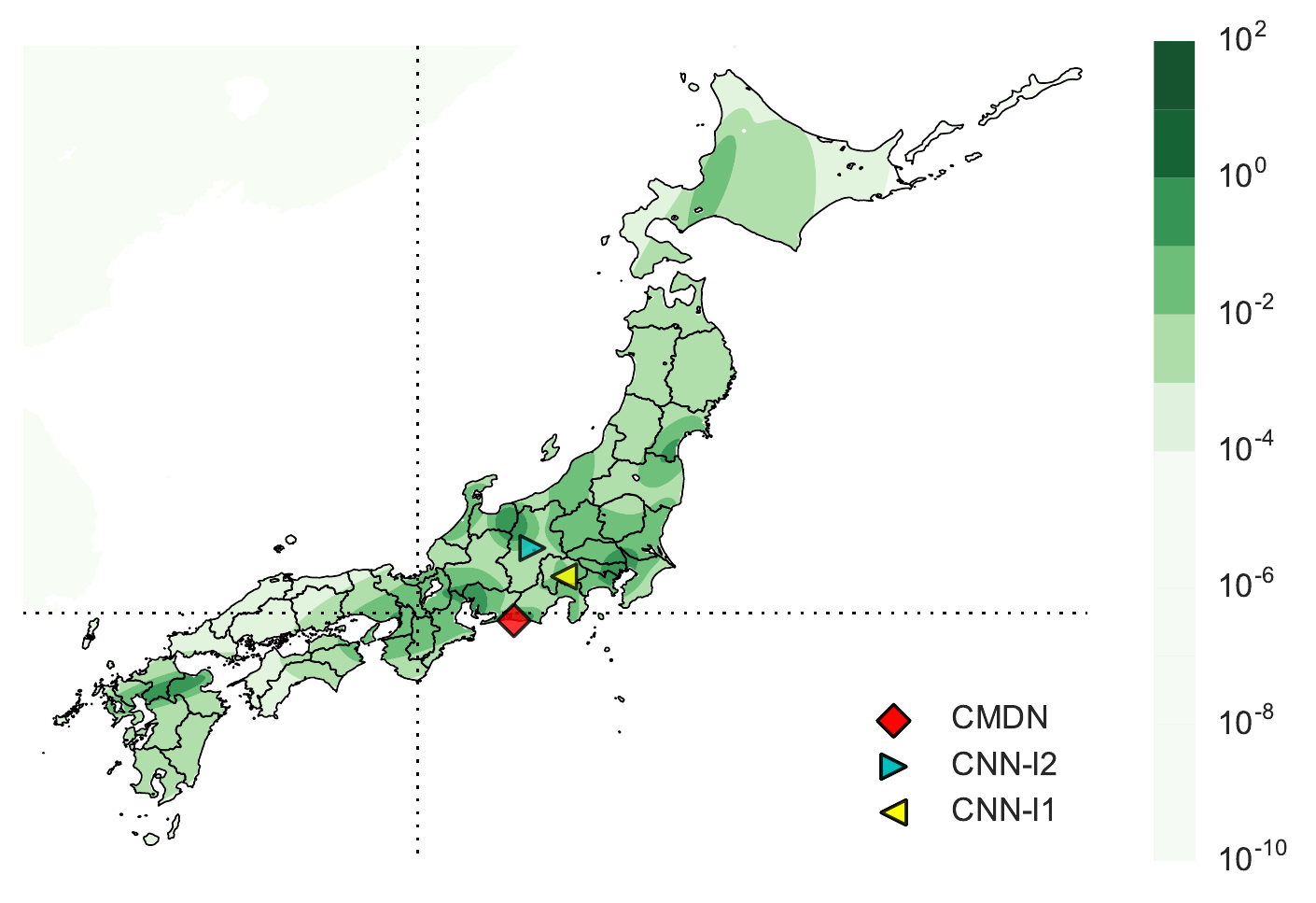}
        \caption{\textit{I've already slept a lot, but I still feel sleepy...} \\ likelihood: 7.0, distance: 174.4 (km)}
        \label{fig:1b}
      \end{subfigure}
      ~ %add desired spacing between images, e.g. ~, \quad, \qquad, and \hfill.
        \begin{subfigure}[b]{0.319\textwidth}
        \includegraphics[width=\textwidth]{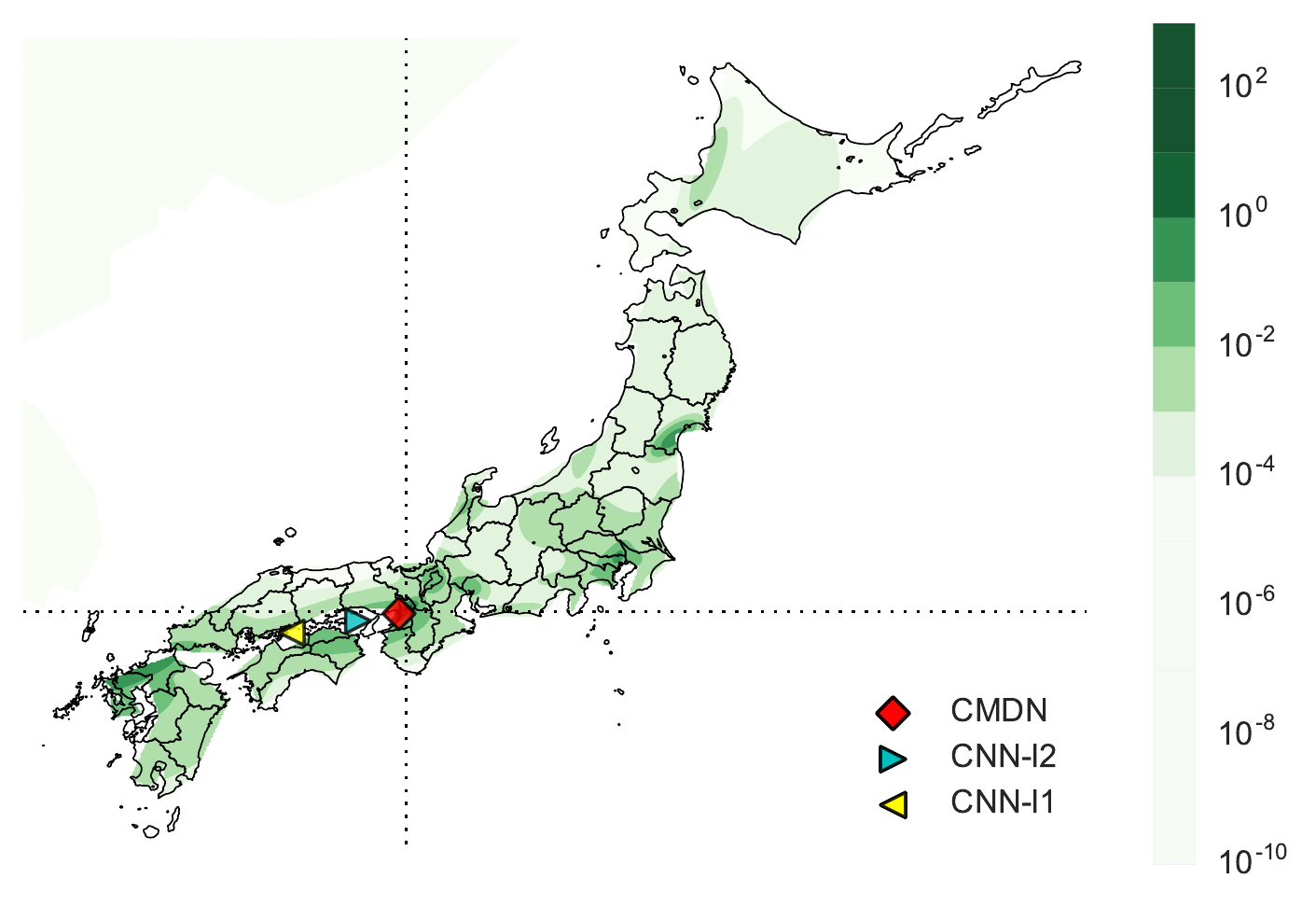}
              \caption{\textit{I will not return to Fukuoka for the first time in Osaka} \\likelihood: 75.6, distance: 13.5 (km)}
               \label{fig:1c}
      \end{subfigure}
      \caption{Original tweet location with the estimated distribution by our proposed method, CMDN. The intersection of dotted line represents the true tweet location. The red diamond represents the estimated location by the CMDN. The blue and yellow triangle represents the CNN for regression with $\ell_2, \ell_1$-loss.}
      \label{fig:examples}
\end{figure*}

\section{Introduction}

Geographic information related to Twitter enriches the availability of data resources. Such information is indispensable for various practical applications such as early earthquake detection \cite{Sakaki2010Earthquake}, infectious disease dispersion assessment \cite{Broniatowski2013health}, and regional user behavior assessment during an election period \cite{Caldarelli2014Election}.
However the application performance depends strongly on the number of geo-tagged tweets, which account for fewer than 0.5 \% of all tweets \cite{Cheng2010iknow}.

To extend the possibilities of the geographic information, a great deal of effort has been devoted to specifying the geolocation automatically \cite{Han2014}.
These studies are classified roughly into two aspects, the \textit{User level} \cite{Cheng2010iknow,Han2014,jurgens2015geolocation,rahimi2015exploiting}  and the \textit{Message level} \cite{dredze2016rm,liu2016where,priedhorsky2014inferring} prediction.
The former predicts the residential area. The latter one predicts the place that the user mentioned.
This study targeted the latter problem, with message level prediction, involving the following three levels of difficulty.

First, most tweets lack information to identify the true geolocation.
In general, many tweets do not include geolocation identifiable words. Therefore, it is difficult even for humans to identify a geolocation (Figure \ref{fig:1b}).

\begin{figure}[t]
	\centering
  \includegraphics[width=0.4\textwidth]{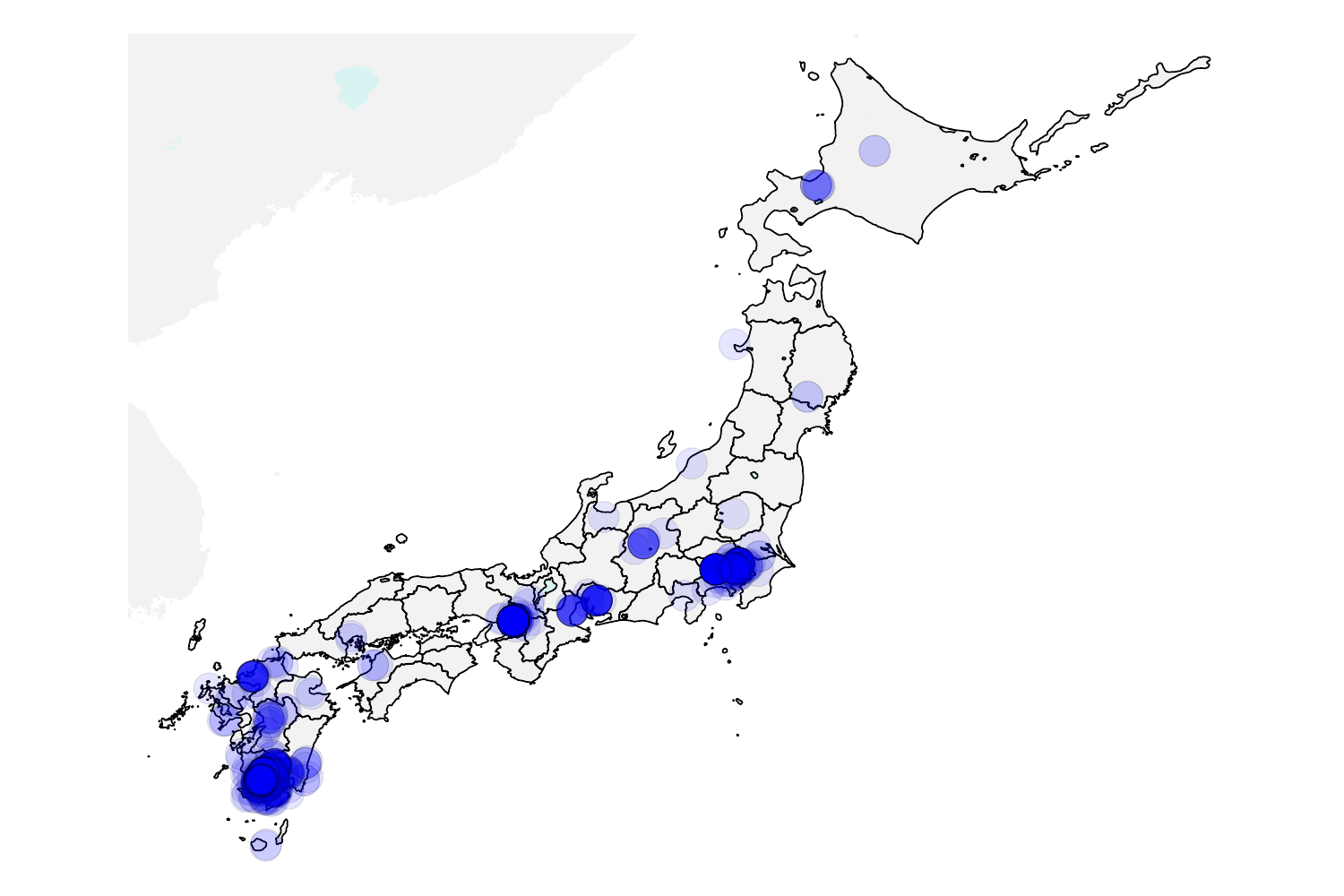}
  \caption{The places where the word ``\textit{Sakurajima}'’ was mentioned in training data are projected on the map. The word ``\textit{Sakurajima}" is mentioned in multiple places because this word represents multiple locations.}
  \label{fig:sakura}
\end{figure}

Next, some location names involve ambiguity because a word refers to multiple locations. For example, places called ``{\it Portland}" exist in several locations worldwide.
Similarly, this ambiguity also arises within a single country, as shown in Figure \ref{fig:sakura}.
Although additional context words are necessary to identify the exact location, many tweets do not include such clue words for identifying the location.
As for such tweet, the real-valued point estimation is expected to be degraded by regression towards the mean \cite{stigler1997regression}.

Finally, if a user states the word that represents the exact location, the user is not necessarily there.
In case the user describes several places in the tweet, contextual comprehension of the tweet is needed to identify the true location (Figure \ref{fig:1c}).

In contrast to most studies, this study was conducted to resolve these issues based on the density estimation approach.
A salient benefit of density estimation is to enable comprehension of the uncertainty related to the tweet user location because it propagates from the estimated distribution and handles tweets distributed to multiple points properly.
Figure \ref{fig:examples} shows each estimated density as a heatmap.
The estimated distribution is concentrated near the true location (Figure \ref{fig:1a}) and vice versa (Figure \ref{fig:1b}) if the tweet includes plenty of clues. Furthermore, the density-based approach can accommodate the representation of multiple output data, whereas the regression-based approach cannot (Figure \ref{fig:1c}).

The density-based approach provides additional benefits for practical application. The estimated density appends the estimation reliability for each tweet as the likelihood value. For reliable estimation, the estimated density provides the high likelihood (Figure \ref{fig:1a} and \ref{fig:1c}) and vice versa (Figure \ref{fig:1b}).

To realize this modeling, we propose a \textbf{Convolutional Mixture Density Network (CMDN)}, a method for estimating the geolocation density estimation from text data. Actually, CMDN extracts valuable features using a convolutional neural network architecture and converts these features to mixture density parameters. Our experimentally obtained results reveal that not merely the high prediction performance, but also the reliability measure works properly for filtering out uncertain estimations.

\section{Related work}

Social media geolocation has been undertaken on various platforms such as Facebook \cite{backstrom2010find}, Flickr \cite{serdyukov2009placing}, and Wikipedia \cite{lieberman2009you}. Especially, Twitter geolocation is the predominant field among all of them because of its availability \cite{Han2014}.

Twitter geolocation methods are not confined to text information. Diverse information can facilitate geolocation performance such as meta information (time \cite{dredze2016rm} and estimated age and gender \cite{pavalanathan2015confounds}), social network structures \cite{jurgens2015geolocation,rahimi2015exploiting}, and user movement \cite{liu2016where}. Many studies, however, have attempted user level geolocation, not the message level. Although the user level geolocation is certainly effective for some applications, message level geolocation supports fine-grained analyses.

However, \newcite{priedhorsky2014inferring} has attempted density estimation for message level geolocation by estimating a word-independent Gaussian Mixture Model and then combining them to derive each tweet density. Although the paper proposed many weight estimation methods, many of them depend strongly on locally distributed words. Our proposed method, CMDN, enables estimate of the geolocation density from the text sequence in the End-to-End manner and considers the marginal context of the tweet via CNN.

\section{Convolutional Neural Network for Regression (Unimodal output)}
\label{sec:cnn_reg}
We start by introducing the convolutional neural network \cite{fukushima1980neocognitron,lecun1998gradient} (CNN) for a regression problem, which directly estimates the real-valued output as our baseline method.
Our regression formulation is almost identical to that of CNN for document classification based on \newcite{kim2014cnn}. We merely remove the softmax layer and replace the loss function.

We assume that a tweet has length $L$ (padded where necessary) and that $w_{i}$ represents the word $i$-th index.
We project the words $\mathbf{w}_i$ to vectors $\mathbf{x}_i$ through an embedding matrix $\mathbf{W}_e$, $\mathbf{x}_{i} = \mathbf{W}_{e}\mathbf{w}_i \in \mathbb{R}^d$, where $d$ represents the size of the embeddings. We compose the sentence vector $\mathbf{x}_{1:L}$ by concatenating the word vectors $\mathbf{x}_i$, as
\begin{align*}
    \mathbf{x}_{1:L} =  \mathbf{x}_1 \oplus \mathbf{x}_2 \oplus \dots \oplus \mathbf{x}_L \in \mathbb{R}^{Ld}.
\end{align*}
In that equation, $\oplus$ represents vector concatenation.

To extract valuable features from sentences, we apply filter matrix $\mathbf{W}_f$ for every part of the sentence vector with window size $\ell$ as
% filter
\begin{align*}
    \mathbf{W}_f &= [\mathbf{w}_{f, 1}, \mathbf{w}_{f, 2}, \dots, \mathbf{w}_{f, m}]^{\top}\\
    c_{i, j} &= \phi(\mathbf{w}_{f, j}^{\top}\mathbf{x}_{i:i+\ell} + b_{f, j})
\end{align*}
where $m$ stands for the number of feature maps, and $\phi$ represents the activation function.
As described herein, we employ ReLU \cite{nair2010rectified} as the activation function.

For each filter's output, we apply 1-max pooling to extract the most probable window feature $\hat{c}_{j}$. Then we compose the abstracted feature vectors $\mathbf{h}$ by concatenating each pooled feature $\hat{c}_{j}$ as shown below.
% pooling & hidden layer
\begin{align*}
    \hat{c}_j &= \max_{i\in\left\{1,\dots, L-\ell+1\right\}} c_{i, j}\\
    \mathbf{h} &= [\hat{c}_1, \hat{c}_2, \dots, \hat{c}_m]^{\top}
\end{align*}

% output
Finally, we estimate the real value output $\mathbf{y}$ using the abstracted feature vector $\mathbf{h}$ as
\begin{align*}
    \hat{\mathbf{y}} = \mathbf{W}_r \mathbf{h} + \mathbf{b}_r \in \mathbb{R}^{q}
\end{align*}
where $q$ signifies the output dimension, $\mathbf{W}_r \in \mathbb{R}^{q\times m}$, denotes the regression weight matrix, and  $\mathbf{b}_r \in \mathbb{R}^q$ represent the bias vectors for regression.

% loss
To optimize the regression model, we use two loss functions between the true value $\mathbf{y}$ and estimated value $\hat{\mathbf{y}}$. One is $\ell_2$-loss as
\begin{align*}
    \sum_{n=1}^N \|\mathbf{y}_n - \hat{\mathbf{y}}_n\|^2_2
\end{align*}
and another is robust loss function for outlier, $\ell_1$-loss,
\begin{align*}
    \sum_{n=1}^N \|\mathbf{y}_n - \hat{\mathbf{y}}_n\|_1 = \sum_{n=1}^N \sum_{q'=1}^q |{y}_{n, q'} - \hat{y}_{n, q'}|
\end{align*}
where $N$ represents the sample size. The $\ell_1$-loss shrinks the outlier effects for estimation rather than the $\ell_2$ one.
% * <s.wakamiya@gmail.com> 2017-02-03T14:26:34.257Z:
%
% > shrink
% shrinks ?
%
% ^ <hyate.iso@gmail.com> 2017-02-04T06:42:55.837Z.

% * <eiji.aramaki@gmail.com> 2017-02-02T07:08:20.197Z:
%
% このモデル がだめなこと ＝  by using text data ともとれる
% This model usually demonstrates the feasibility, but for the task of geolocation estimation, it suffers from the geolocation specific problems, such as location ambiguity,  outliers....
%
% ^
% ^ <hyate.iso@gmail.com> 2017-02-03T01:21:40.633Z:
%
% 3章は人によっては完全に読み飛ばせるように，欠点などは4章の頭に追加して提案モデルの導入にしました．
%
% ^ <hyate.iso@gmail.com> 2017-02-04T06:42:58.811Z.

% * <eiji.aramaki@gmail.com> 2017-02-02T07:11:40.871Z:
%
% motivating our mixture density approach mentioned in the next section,.
%
% ^ <hyate.iso@gmail.com> 2017-02-04T06:43:00.699Z.

\section{Convolutional Mixture Density Network (Multimodal output)}
% * <hyate.iso@gmail.com> 2017-02-02T06:49:09.009Z:
%
% > Convolutional Mixture Density Network (Multimodal output)
% Subsection を作って，見出しだけ見れば手法を追えるようにする
%
% ^ <hyate.iso@gmail.com> 2017-02-04T06:43:03.191Z.
In the previous section, we introduced the CNN method for regression problem.
This regression formulation is good for addressing well-defined problems. Both the input and output have one-to-one correspondence. Our tweet corpus is fundamentally unfulfilled with this assumption.
Therefore, a more flexible model must be used to represent the richer information.

In this section, we propose a novel architecture for text to density estimation, \textbf{Convolutional Mixture Density Network (CMDN)}, which is the extension of Mixture Density Network by \newcite{bishop1994mdn}. In contrast to the regression approach that directly represents the output values $\hat{\mathbf{y}}$, CMDN can accommodate more complex information as the probability distribution.
The CMDN estimates the parameters of Gaussian mixture model $\pi_k, \bm{\mu}_k, \bm{\Sigma}_k$, where $k$ is the $k$-th mixture component using the same abstracted features $\mathbf{h}$ in the CNN regression (Sec \ref{sec:cnn_reg}).

\subsection{Parameter estimation by neural network}
Presuming that density consists of $K$ components of multivariate normal distribution $\mathcal{N}(\mathbf{y}| \bm{\mu}_k, \bm{\Sigma}_k)$, then the number of each $q$-dimensional normal distribution parameters are $p=\frac{q(q+3)}{2}$ ($q$ parameters for each mean $\bm{\mu}_k$ and diagonal values of the covariance matrix $\bm{\Sigma}_k$ and $\frac{q(q-1)}{2}$ parameters for correlation parameters $\rho$) between dimension outputs.

For $q=2$, each component of the parameters is represented as
\begin{align*}
        \bm{\mu}_k = \begin{pmatrix} \mu_{k,1} \\ \mu_{k, 2} \end{pmatrix},
    \bm{\Sigma}_k = \begin{pmatrix} \sigma_{k, 1}^2 & \rho_k \sigma_{k,1} \sigma_{k,2} \\
                             \rho_k \sigma_{k,1} \sigma_{k,2}  & \sigma_{k,2}^2 \end{pmatrix}.
\end{align*}

To estimate these parameters, we first project the hidden layer $\mathbf{h}$ into the required number of parameters space $\bm{\theta}$ as
% parameter
\begin{align*}
    \bm{\theta} = \mathbf{W}_p \mathbf{h} + \mathbf{b}_p \in \mathbb{R}^{Kp},
\end{align*}
where $\mathbf{W}_p \in \mathbb{R}^{Kp\times m}$ and $\mathbf{b}_p \in \mathbb{R}^{Kp}$ respectively denote the weight matrix and bias vector.

Although the parameter $\bm{\theta}$ has a sufficient number of parameters to represent the $K$ mixture model, these parameters are not optimal for insertion to the parameters of the multivariate normal distribution $\mathcal{N}(\mathbf{y}|\bm{\mu}_k, \bm{\Sigma}_k)$, and its mixture weight $\pi_k$.

The mixture weights $\pi_k$ must be positive values and must sum to 1. The variance parameters $\sigma$ must be a positive real value. The correlation parameter $\rho_k$ must be $(-1, 1)$.
For this purpose, we transform real-valued outputs $\bm{\theta}$ into the optimal range for each parameter of mixture density.

\subsection{Parameter conversion}
% parameter seed
For simplicity, we first decompose $\bm{\theta}$ to each mixture parameter $\bm{\theta}_k$ as
\begin{align*}
	\bm{\theta} &= \bm{\theta}_1 \oplus \bm{\theta}_2 \oplus \dots \oplus \bm{\theta}_K\\
    \bm{\theta}_k &= (\theta_{\pi_k}, \theta_{\mu_{k, 1}},\theta_{\mu_{k, 2}}, \theta_{\sigma_{k, 1}}, \theta_{\sigma_{k, 2}}, \theta_{\rho_{k}}).
\end{align*}
To restrict each parameter range, we convert each vanilla parameter $\bm{\theta}_k$ as
% parameters
\begin{align*}
    \pi_k & = \mbox{softmax}(\theta_{\pi_k})\\
    &= \frac{\exp(\theta_{\pi_k})}{\sum_{k'=1}^K\exp(\theta_{\pi_k'})} \in (0, 1)\\
    \mu_{k,j} &= \theta_{\mu_{k, j}} \in \mathbb{R},\\
    \sigma_{k,j} &= \mbox{softplus}(\theta_{\sigma_{k, j}})\\
    &= \ln\left(1+\exp(\theta_{\sigma_{k, j}})\right) \in (0, \infty),\\
    \rho_{k} & = \mbox{softsign}(\theta_{\rho_k})\\
    &= \frac {\theta_{\rho_k}}{1+|\theta_{\rho_k}|} \in (-1, 1).
\end{align*}

The original MDN paper \cite{bishop1994mdn} and its well known application of MDN for handwriting generation \cite{graves2013generating} used an exponential function for transforming variance parameter $\sigma$ and a hyperbolic tangent function for the correlation parameter $\rho$. However, we use \textit{softplus} \cite{glorot2011deep} for variance and \textit{softsign} \cite{glorot2010understanding} for correlation.

Replacing the activation function in the output layer prevents these gradient problems. Actually, these gradient values are often exploded or nonexistent. Our proposed transformation is effective to achieve rapid convergence and stable learning.

\subsection{Loss function for parameter estimation}
To optimize the mixture density model, we use negative log likelihood as the training loss:
% negative log likelihood
\begin{align*}
    -\sum_{n=1}^N \ln\left( \sum_{k=1}^K \pi_k \mathcal{N}(\mathbf{y}_n|\bm{\mu}_k, \bm{\Sigma}_k)\right).
\end{align*}

\section{Experiments}
In this section, we formalize our problem setting and clarify our proposed model effectiveness.

\subsection{Problem setting}
This study explores CMDN performance from two perspectives.

Our first experiment is to predict the geographic coordinates by which each user stated using the only single tweet content. We evaluate the mean and median value of distances measured using Vincenty's formula \cite{vincenty1975direct} between the estimated and true geographic coordinates for the overall dataset.
% * <s.wakamiya@gmail.com> 2017-02-03T14:33:56.444Z:
%
% > geographic coordinate
% 緯度経度をいう場合，coordinateは複数形になるのでsつけました
%
% ^ <hyate.iso@gmail.com> 2017-02-04T06:43:15.154Z.

The second experiment is used to filter out the unreliable estimation quantitatively using the likelihood-based threshold. Each estimated density assigns the likelihood value for every point of the location. We designate the likelihood values as reliability indicators for the respective estimated locations. Then, we remove the estimation from the lowest likelihood value and calculate both the mean and median values. This indicator is filtered correctly out the unreliable estimations if the statistics decrease monotonically.

\subsubsection*{Location estimation by estimated density}
In contrast to the regression approach, it is necessary to specify the estimation point from the estimated density of each tweet. In accordance with \newcite{bishop1994mdn}, we employ the mode value of estimated density as the estimated location $\hat{\mathbf{y}}$. The mode value of the probability distribution can be found by numerical optimization, but it requires overly high costs for scalable estimation. For a simple and scalable approximation for seeking the mode value of the estimated density, we restrict the search space to each mean value of the mixture components as shown below:
\begin{align*}
    \hat{\mathbf{y}} = \argmax_{\mathbf{y} \in \{\bm{\mu}_1, \dots,\bm{\mu}_K\}} \sum_{k=1}^K \pi_k \mathcal{N}(\mathbf{y}|\bm{\mu}_k, \bm{\Sigma}_k).
\end{align*}

\subsection{Dataset}

Our tweet corpus consists of 24,633,478 Japanese tweets posted from July 14, 2011 to July 31, 2012. Our corpus statistics are presented in Table \ref{tab:data}. We split our corpus randomly into training for 20M tweets, development for 2M, and test for 2M.
% * <s.wakamiya@gmail.com> 2017-02-03T14:36:52.601Z:
%
% > was consists of
% consists of　にしました
%
% ^ <hyate.iso@gmail.com> 2017-02-04T06:43:18.643Z.

\begin{table}
% * <s.wakamiya@gmail.com> 2017-02-03T10:15:27.359Z:
%
% 数値なので右寄せにしました
%
% ^ <hyate.iso@gmail.com> 2017-02-04T06:43:23.404Z.
\centering
\begin{tabular}{lr}
  Dataset      & \\
  \midrule
  \# of tweets & 24,633,478 \\
  \# of users  & 276,248\\
  Average \# of word & 16.0 \\
  \# of vocabulary & 351,752
\end{tabular}
\caption{Dataset statistics.}
\label{tab:data}
\end{table}

\subsection{Comparative models}
We compare our proposed model effectiveness by controlling experiment procedures, which replace the model components one-by-one. We also provide simple baseline performance. The following model configurations are presented in Table \ref{tab:config}.

\textbf{Mean:} Mean value of the training data locations.

\textbf{Median:} Median value of the training data locations.

\textbf{Enet:} Elastic Net regression \cite{zou2005regularization}, which consists of ordinary least squares regression with $\ell_2, \ell_1$-regularization.

\textbf{MLP-l2:} Multi Layer Perceptron with $\ell_2$-loss \cite{minsky69perceptrons}

\textbf{MLP-l1:} Multi Layer Perceptron with $\ell_1$-loss.

\textbf{CNN-l2:} Convolutional Neural Network for regression with $\ell_2$-loss based on \newcite{kim2014cnn}

\textbf{CNN-l1:} Convolutional Neural Network for regression with $\ell_1$-loss.

\textbf{MDN:} Mode value of Mixture Density Network \cite{bishop1994mdn}

\textbf{CMDN:} Mode value of Convolutional Mixture Density Network (Proposed)

\begin{table}
\centering
\begin{tabular}{lr}
% * <s.wakamiya@gmail.com> 2017-02-03T10:15:20.643Z:
%
% 数値なので右寄せにしました
%
% ^ <hyate.iso@gmail.com> 2017-02-04T06:43:27.205Z.
  Parameters      & \\
  \midrule
  \# of mixture         & 50 \\
  Embedding dimension   & 300 \\
  Window sizes          & 3, 4, 5\\
  Each filter size      & 128\\
  Dropout rate \shortcite{srivastava2014dropout}& 0.2\\
  Batch size            & 500 \\
  Learning rate         & 0.0001\\
  Optimization          & Adam \shortcite{kingma2014adam}
\end{tabular}
\caption{Model configurations.}
\label{tab:config}
\end{table}

\subsection{Results}

\subsubsection{Geolocation performance}
The experimental geolocation performance results are presented in Table \ref{tab:geolocation}.

Overall results show that our proposed model \textbf{CMDN} provides the lowest median error distance: \textbf{CNN-l1} is the lowest mean error distance. Also, CMDN gives similar mean error distances to those of CNN-l1. Both CMDN and CNN-l1 outperform all others in comparative models.

In addition to the results obtained for the feature extraction part, the CNN-based model consistently achieved better prediction performance than the vanilla MLP-based model measured by both the mean and median.

\begin{table}
\centering
\begin{tabular}{lrr}
% * <s.wakamiya@gmail.com> 2017-02-03T10:13:04.780Z:
%
% 数値なので右寄せにしました
%
% ^ <hyate.iso@gmail.com> 2017-02-04T06:35:46.781Z.
  Method          & Mean (km) & Median (km)\\
  \midrule
  \textbf{CMDN}       & 159.4     & \textbf{10.7}  \\
  \textbf{CNN-l1}     & \textbf{147.5}     & 28.2 \\
  \textbf{CNN-l2}     & 166.5     & 80.5 \\
  \textbf{MDN}        & 251.4     & 92.9  \\
  \textbf{MLP-l1}     & 224.0     & 75.0  \\
  \textbf{MLP-l2}     & 226.8     & 140.3 \\
  \textbf{Enet}       & 197.2     & 144.6 \\
  \textbf{Mean}       & 279.4     & 173.9 \\
  \textbf{Median}     & 253.7     & 96.1
\end{tabular}
\caption{The prediction performance. The lower value is the better estimation.}
\label{tab:geolocation}
\end{table}

\subsubsection{Likelihood based threshold}
\begin{figure}[t]
  \includegraphics[width=0.5\textwidth]{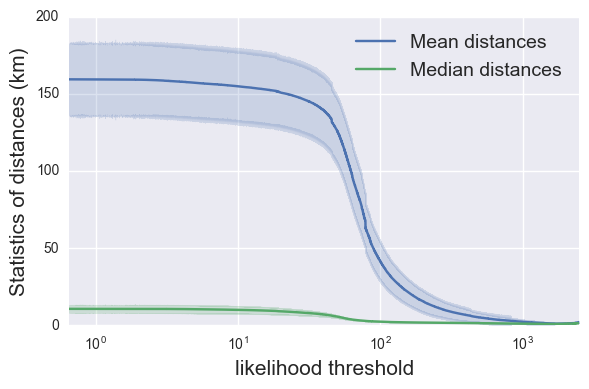} % あとで対応
  \caption{Likelihood-based threshold results with 95\% confidence interval for our proposed model: CMDN. To extract reliable (lower value of distance measures) geolocated tweets, we can increase the lower bound of likelihood.  The standard deviations of both the mean and median are calculated using Bootstrap methods \cite{efron1979bootstrap}.}
  \label{fig:threshold_stats}
\end{figure}

\begin{figure}[t]
  \includegraphics[width=0.5\textwidth]{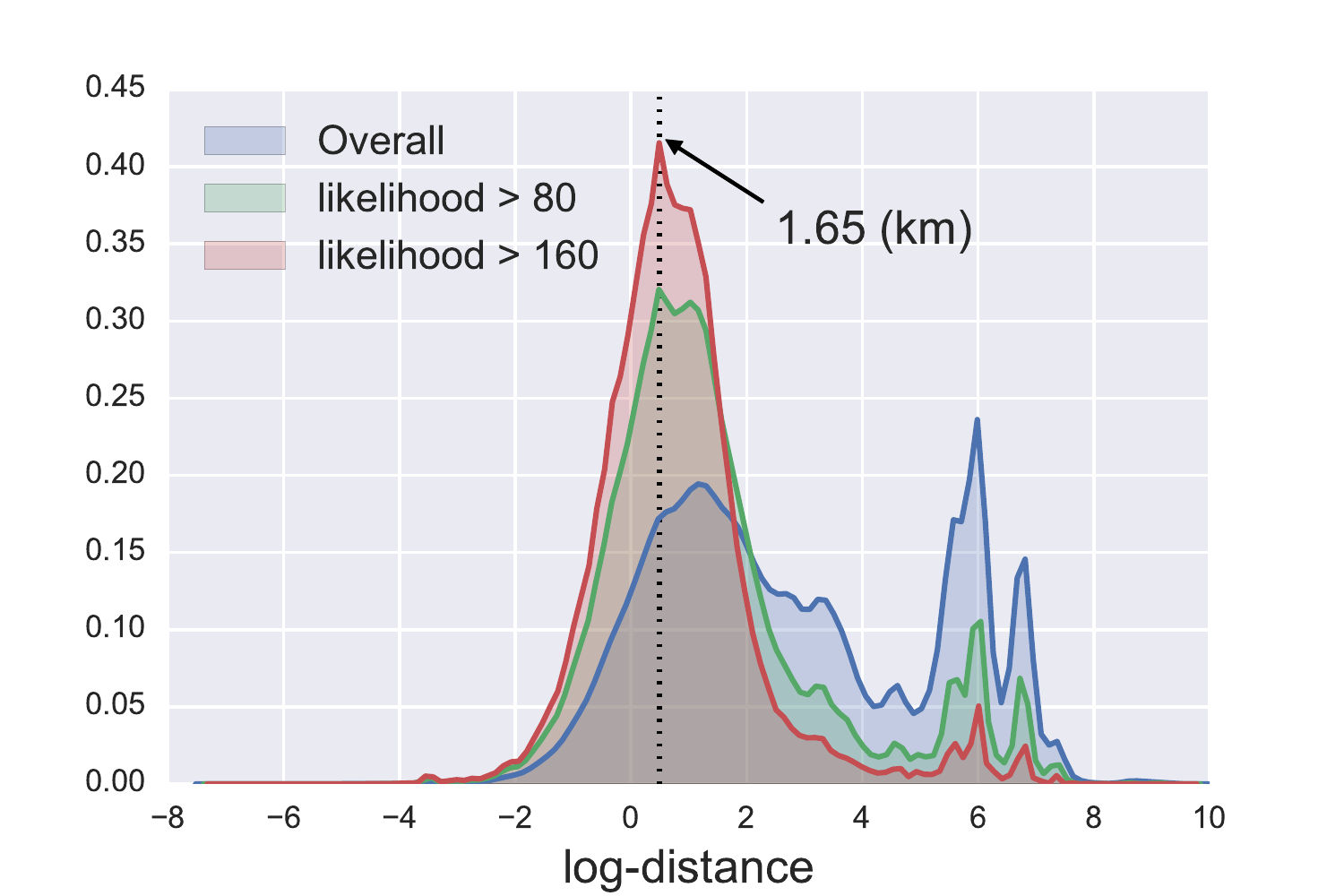}
  \caption{Output distribution with several likelihood thresholds. The small log-distance represents the better prediction.}
  \label{fig:threshold_dists}
\end{figure}

We show how the likelihood-based threshold affects both mean and median statistics in Figure \ref{fig:threshold_stats}.
The likelihood-based threshold consistently decreases both statistics in proportion to the likelihood lower bound increases. Especially, these statistics dramatically decrease when the likelihood is between $10^1$ and $10^2$. Several likelihood bounds results are presented in Figure \ref{fig:threshold_dists}. Although the model mistakenly estimates many incorrect predictions for the overall dataset (blue), the likelihood base threshold correctly prunes the outlier estimations.

\section{Discussion}
Our proposed model provides high accuracy for our experimental data.
Moreover, likelihood-based thresholds reveal a consistent indicator of estimation certainty. In this section, we further explore the properties of the CMDN from the viewpoint of difference of the loss function.

\subsection*{Loss function difference}
\label{sec:loss_differences}
\begin{figure}[t]
  \includegraphics[width=0.5\textwidth]{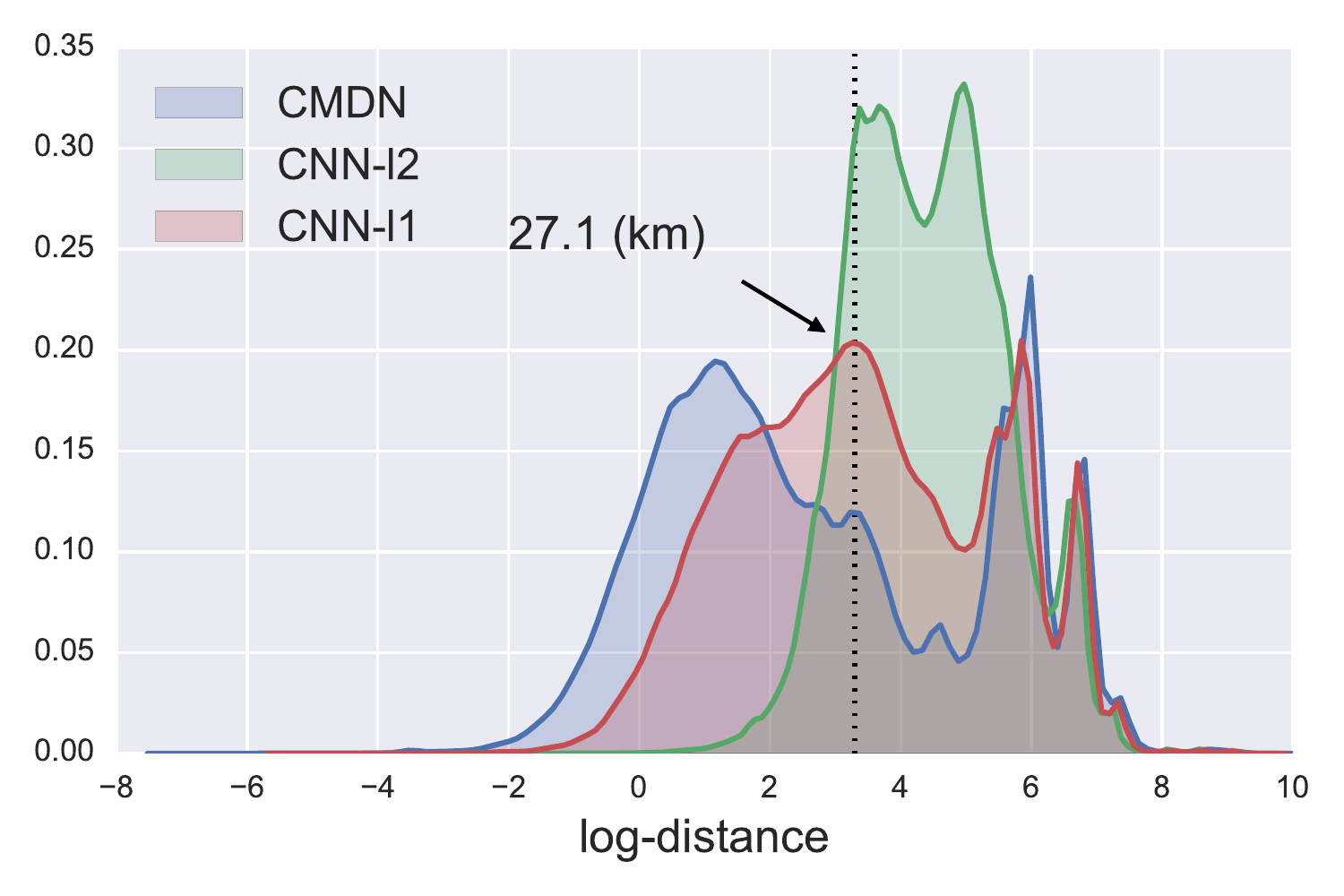}
  \caption{Loss function difference: Lower values represent better predictions. Our proposed model, CMDN (blue), tends to be the most aggressive estimation.}
  \label{fig:loss_differences}
\end{figure}

    We compare the CNN-based model with different loss functions, $\ell_2, \ell_1$-loss for regression \textbf{CNN-l2, CNN-l1} and negative log-likelihood for mixture density model \textbf{CMDN}. The output distance distributions are shown in Figure \ref{fig:loss_differences}.

   Although $\ell_2$-loss denotes the worst performance measured by mean and median among CNN-based models (Table \ref{tab:geolocation}), it represents the lowest outlier ratio. Consequently, the $\ell_2$-loss is the most conservative estimate for our skewed corpus.

    We can infer that the difference between the competitive model CNN-l1 and our proposed model CMDN is their estimation aggressiveness. The median of CMDN is remarkably lower than of CNN-l1, but the mean of CMDN is slightly larger than that of CNN-l1. The reason for these phenomena is the difference of the loss functions behavior for multiple candidate data. Even though the $\ell_1$-loss function is robust to outliers, the estimation deteriorates when the candidates appear with similar possibilities. In contrast, CMDN can accommodate multiple candidates as multiple mixture components. Therefore, CMDN imposes the appropriate probabilities for several candidates and picks up the most probable point as the prediction. In short, CMDN's predictions become more aggressive than CNN-l1's. Consequently, CMDN's median value becomes lower than CNN-l1's.

    An important shortcoming of aggressive prediction is that the estimation deteriorates when the estimation fails. The mean value tends to be affected strongly by the outlier estimation. Therefore, CMDN's mean value becomes higher than that of CNN-l1's.

    However, CMDN can overcome this shortcoming using a likelihood-based threshold, which first filters out the outlier. Therefore, we conclude that the negative log-likelihood for mixture density is better than those of other loss functions $\ell_2$ and $\ell_1$-loss for the regression.

\section{Future work}
This study assessed the performance of our proposed model, \textbf{CMDN}, using text data alone. Although text-only estimation is readily applicable to existing resources, we still have room for improvement of the prediction performance. The winner of the Twitter Geolocation Prediction Shared Task, \newcite{miura2016simple}, proposed that the unified architecture handle several meta-data such as the user location, user description, and time zone for predicting geolocations. The CMDN can integrate this information in the same manner.
% * <s.wakamiya@gmail.com> 2017-02-03T10:28:22.710Z:
%
% > our proposed model, \textbf{CMDN}, performance by
% We investigate the performance of our proposed model, \textbf{CMDN}, by
%
% ^ <hyate.iso@gmail.com> 2017-02-04T06:37:04.948Z.
% * <s.wakamiya@gmail.com> 2017-02-03T10:20:03.823Z:
%
% > we still have the room for improvements the prediction performance.
% -> we still have a room for improving the prediction performance.
%
% ^ <hyate.iso@gmail.com> 2017-02-04T06:37:23.023Z.

Furthermore, \newcite{liu2016where} reports that the user home location strongly affects location prediction for a single tweet. For example, a routine tweet is fundamentally unpredictable using text contents alone, but if the user home location is known, this information is a valuable indication for evaluating the tweet. As future work, we plan to develop a unified architecture that incorporates user movement information using a recurrent neural network.

In contrast, our objective function might be no longer useful for world scale geolocation because ours approximates the spherical coordinates into the real coordinate space. This approximation error tends to become larger for the larger scale geolocation inference.
We will explore our method's geolocation performance using the world scale geolocation dataset such as W-NUT data~\cite{han2016twitter}.

\section{Conclusion}
This study clarified the capabilities of the density estimation approach to Twitter geolocation.
Our proposed model, CMDN, performed not only with high accuracy for our experimental data; it also extracted reliable geolocated tweets using likelihood-based thresholds. Results show that CMDN merely requires the tweet message contents to identify its geolocation, while obviating preparation of meta-information. Consequently, CMDN can contribute to extension of the fields in which geographic information application can be used.

%\bibliography{acl2017}
\bibliography{acl2017}
\bibliographystyle{acl_natbib}

\end{document}